\documentclass{article}

\usepackage{lmodern}
\usepackage{PRIMEarxiv}

\usepackage[utf8]{inputenc}
\usepackage[T1]{fontenc}
\usepackage{amsmath,amsfonts,amssymb}
\usepackage{booktabs}
\usepackage{array}
\usepackage{graphicx}
\usepackage{xcolor}
\usepackage{microtype}
\usepackage{fancyhdr}
\usepackage[colorlinks=true,linkcolor=blue,citecolor=blue,urlcolor=blue,bookmarks=true]{hyperref}
\usepackage{url}
\usepackage[figure,table]{hypcap}
\graphicspath{{figures/}}

\newcommand{\method}{SAR-TopoGuard}
\newcommand{\doiurl}{https://doi.org/10.1109/LGRS.2026.3753492}

\title{SAR-TopoGuard: Training-Free, Speckle-Aware Topological Verification for Structure-Preserving SAR-to-Optical Translation
\thanks{\textit{\underline{Citation}}: S.~Sarkar, D.~Ghosh, S.~Biswas, K.~Mandal, and D.~Sobya, ``SAR-TopoGuard: Training-free, speckle-aware topological verification for structure-preserving SAR-to-optical translation,'' \textit{IEEE Geoscience and Remote Sensing Letters}, 2026, doi: \href{\doiurl}{10.1109/LGRS.2026.3753492}.\newline
\copyright~2026 IEEE. Personal use of this material is permitted. Permission from IEEE must be obtained for all other uses, in any current or future media, including reprinting/republishing this material for advertising or promotional purposes, creating new collective works, for resale or redistribution to servers or lists, or reuse of any copyrighted component of this work in other works.\newline
This is the authors' accepted manuscript. The published version is available on IEEE Xplore: \url{https://ieeexplore.ieee.org/document/11793632}.}}

\author{
Satabarto Sarkar\textsuperscript{1,*,\textdagger}, Debojyoti Ghosh\textsuperscript{1,\textdagger}, Subhranil Biswas\textsuperscript{1}, Kaustav Mandal\textsuperscript{1}, and D.~Sobya\textsuperscript{1}\\[0.5ex]
{\normalfont \textsuperscript{1}Department of Computer Science and Engineering, Institute of Engineering and Management,}\\
{\normalfont University of Engineering and Management, Kolkata 700091, India}\\[0.5ex]
{\normalfont\small \textsuperscript{*}Corresponding author: \href{mailto:satabarto.sarkar2023@iem.edu.in}{\texttt{satabarto.sarkar2023@iem.edu.in}}\quad \textsuperscript{\textdagger}These authors contributed equally to this work.}
}

\begin{document}
\maketitle

\begin{abstract}
SAR-to-optical translators often synthesize detail that the radar measurement does not support, and pixel metrics alone cannot reveal it. We propose \method, a training-free inference-time layer for frozen generators: a stochastic sampler draws $K$ candidates (Monte Carlo dropout for GANs, seed variance for diffusion), a speckle-aware verifier ranks them by gradient-magnitude correlation with the SAR input and a hallucination-frequency index, and a SAR-guided frequency gate admits high-frequency detail only where SAR edges support it. With all five hyperparameters frozen and three training seeds, we evaluate on 500 scene-disjoint SEN1-2 test patches, 500 patches of a held-out summer scene, and 200 cross-dataset SEN12MS winter pairs. The layer improves Pix2Pix by $+0.060$, $+0.012$, and $+0.077$~dB PSNR and by $+0.0034$, $+0.0057$, and $+0.0135$ SSIM, respectively (Wilcoxon $p<10^{-6}$), and it improves CycleGAN and CUT on all three sets ($+0.013$ to $+0.076$~dB). It raises edge precision at the cost of edge recall (edge F1 $-1.3\%$ on the test set). On the latent diffusion model COP-GEN, it adds $+0.33$~dB (base) and $+0.16$~dB (Beta) on the winter set, the only set with dual-polarization SAR. An ablation shows that the gate produces the gain for MC-dropout GANs, whose candidates are nearly identical, whereas verifier selection contributes up to $+0.26$~dB for diffusion, whose samples differ. The layer has zero trainable parameters and adds 0.4~ms per patch beyond the $K=5$ generator passes.
\end{abstract}


\section{Introduction}
Synthetic aperture radar (SAR) images the Earth day and night and through cloud, as Moreira et al. detail \cite{moreira2013tutorial}, whereas optical imagery is easier to interpret but fails under cloud. Translating SAR into optical-like images has therefore become an active topic, supported by paired benchmarks such as SEN1-2 of Schmitt et al. \cite{schmitt2018sen12} and conditional GANs following Isola et al. \cite{isola2017image}.

Translation is ill-posed: many optical appearances are consistent with one SAR observation, so generators trained with pixel or adversarial losses produce plausible high-frequency detail, such as field patterns, road-like strokes, or water texture, that the radar does not support. We call such invented, removed, or displaced structure \emph{topological hallucination}. It matters when translated images feed registration and change detection \cite{zitova2003image}, yet pixel scores such as SSIM \cite{wang2004image} average error against one reference and cannot tell whether a detail is supported by the input.

Most countermeasures change the translator or its training (Section~\ref{sec:related}). When a generator is frozen or deployed, the remaining question is whether each synthesized detail has support in the SAR input. MC dropout, shown by Gal and Ghahramani to approximate Bayesian inference \cite{gal2016dropout}, offers several outputs per input, but it has not been used to select or filter translations against the physics of the conditioning modality.

\begin{figure}[t]
\centering
\includegraphics[width=0.92\linewidth]{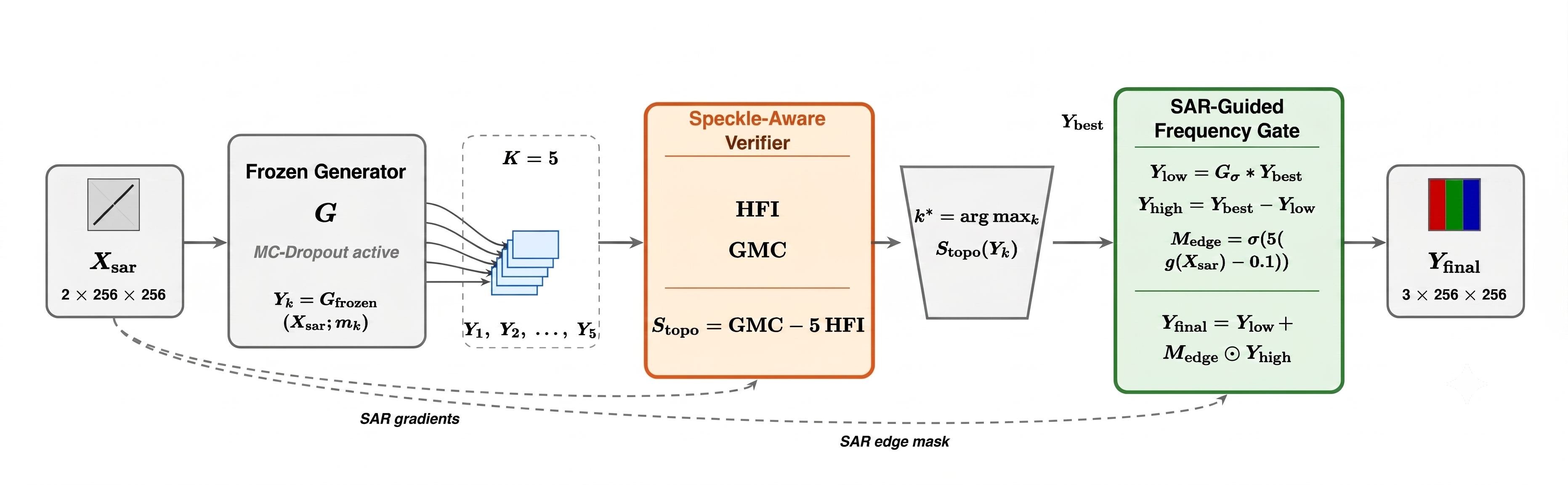}
\caption{Overview of \method. A frozen generator produces $K=5$ candidates from the SAR input (MC dropout for GANs, seed variance for diffusion). The speckle-aware verifier ranks them by gradient-magnitude correlation (GMC) and the hallucination-frequency index (HFI); the selected candidate passes through a SAR-guided frequency gate.}
\label{fig:pipeline}
\end{figure}

We propose \method, a training-free layer that samples $K$ outputs from a frozen generator, selects one with a SAR-conditioned score, and gates its high-frequency residual by SAR edge support (Fig.~\ref{fig:pipeline}). The components (MC dropout, gradient statistics, blur-residual decomposition) are standard; the contributions are:
\begin{enumerate}
\item a SAR-conditioned selection score and gate that need only a stochastic sampler;
\item an evaluation with three training seeds on a scene-disjoint split, a held-out season, and an independent cross-dataset set, across three GANs and two diffusion checkpoints;
\item an ablation that explains when each component helps: the gate improves fidelity when candidates are nearly identical, as for MC-dropout GANs, and selection adds value when samples are diverse, as for diffusion.
\end{enumerate}

\section{Related Work}\label{sec:related}
\paragraph{Translators.} GAN-based SAR-to-optical translation builds on paired Pix2Pix~\cite{isola2017image} and unpaired CycleGAN~\cite{zhu2017unpaired} and CUT~\cite{park2020contrastive}; Zhang et al. add downstream-task guidance~\cite{zhang2025segcyclegan}. Physics-inspired networks derive the generator from numerical schemes: S2O-NPDE uses central differences and Perona--Malik diffusion to suppress speckle while keeping geometry~\cite{zhang2022s2onpde}, and S2O-TDN derives its blocks from thermodynamic finite differences~\cite{zhang2024s2otdn}. Diffusion models condition the denoising chain on SAR~\cite{bai2024conddiff,rombach2022high}, reduce it to one step~\cite{qin2025e3diff}, or model several Copernicus modalities jointly~\cite{espinosa2026copgen,espinosa2025copgenbeta}. These methods improve fidelity by changing the network or its loss, which requires retraining, and none checks individual outputs after deployment.

\paragraph{Position of this work.} \method\ leaves the translator frozen and selects and gates its outputs at inference, so it complements any translator that can produce several samples per input.

\section{Data and Method}
\subsection{Data, Splits, and Preprocessing}\label{sec:data}
We use a 16,000-pair terrain-stratified subset of SEN1-2 \cite{schmitt2018sen12}: co-registered $256\times256$ patches at 10~m (about 2.56~km per side) of Sentinel-1 IW GRD VV backscatter (dB clipped to $[-25,0]$, 8-bit) and Sentinel-2 RGB (B4/B3/B2), from fall (29 scenes) and summer (one scene) acquisitions. Each scene (one Sentinel-1/2 image pair) is assigned to exactly one split: 5,000 training pairs (20 fall scenes), a 250-patch tuning set (3 scenes), and 500 test patches (6 scenes). Two out-of-distribution (OOD) sets are used for neither training nor tuning: 500 patches of the held-out summer scene and 200 winter pairs from SEN12MS \cite{schmitt2019sen12ms} (31 scenes, GeoTIFF with VV and VH in dB and 13-band Sentinel-2). We verified that no scene or patch appears in two splits (Table~\ref{tab:split}).

SAR intensities are heavy-tailed because of multiplicative speckle \cite{moreira2013tutorial}, so both modalities are stretched to $[0,1]$ after log compression of the SAR:
\begin{equation}
 I'=\operatorname{clip}\!\left(\frac{J-P_2}{P_{98}-P_2},0,1\right),\quad J\in\{I_{\mathrm{opt}},\,\log_{10} I_{\mathrm{sar}}\},
\label{eq:normalization}
\end{equation}
where $P_2$ and $P_{98}$ are training-set percentiles. SEN1-2 already stores SAR in dB, so we invert its 8-bit quantization to obtain $\log_{10}I_{\mathrm{sar}}$ and apply no second logarithm. The winter set has 16-bit optical bands, so its percentiles are computed on that set; the GANs receive its VV channel only, as in training.

\begin{table}[t]
\caption{Scene-disjoint splits.}
\label{tab:split}
\centering
\small
\begin{tabular}{@{}lccl@{}}
\toprule
Split & Scenes & Pairs & Use\\
\midrule
Train (fall) & 20 & 5,000 & training of all GANs\\
Tune (fall) & 3 & 250 & reported only (Table~\ref{tab:p2p})\\
Test (fall) & 6 & 500 & main evaluation\\
Summer & 1 & 500 & held-out season (OOD)\\
Winter (SEN12MS) & 31 & 200 & cross-dataset set (OOD)\\
\bottomrule
\end{tabular}
\end{table}

\subsection{Generators and Candidate Sampling}
Pix2Pix (B1) \cite{isola2017image} uses a U-Net generator and PatchGAN discriminator with $L_1$ and adversarial losses; B2 adds a Sobel edge loss ($\lambda_e=10$), a typical training-time structural constraint. Both are trained for 50 epochs (batch 8, Adam, learning rate $2\times10^{-4}$, $\beta=(0.5,0.999)$). CycleGAN \cite{zhu2017unpaired} and CUT \cite{park2020contrastive} use a ResNet-9 generator with dropout in the residual blocks and are trained for 30 epochs. Every GAN is trained from scratch with seeds 0, 1, and 2 on the same split; VV fills both SAR input channels. At inference, the weights are frozen and dropout ($p=0.5$) is re-enabled (the three innermost U-Net decoder blocks, or the residual blocks), so $K$ passes give the candidates
\begin{equation}
 Y_k=G_{\mathrm{frozen}}(X_{\mathrm{sar}};m_k),\qquad k=1,\ldots,K,
 \label{eq:mc}
\end{equation}
with independent dropout masks $m_k$. For the COP-GEN latent diffusion transformers (base and Beta checkpoints) \cite{espinosa2026copgen,espinosa2025copgenbeta}, which are used as released, the candidates are $K$ samples with different initial noise (50 DPM-Solver++ steps), conditioned on Sentinel-1 VV and VH. The sampler is the only backbone-specific part; the verifier and gate are identical for all backbones. Full training settings are listed in Appendix~\ref{app:repro}.

\subsection{Speckle-Aware Topological Verifier}
The verifier treats the SAR input as shared evidence. With Sobel derivatives $D_x$, $D_y$, the channel-mean gradient magnitude and its $r\times r$ average-pooled version ($r=3$) are
\begin{equation}
 g(I)=\frac{1}{C}\sum_{c=1}^{C}\sqrt{(D_x\ast I_c)^2+(D_y\ast I_c)^2},\qquad
 \tilde g(I)=\operatorname{Pool}_r(g(I)).
 \label{eq:gradient}
\end{equation}
A flatness mask marks regions where the SAR shows little structure,
\begin{equation}
 M_{\mathrm{flat}}=\exp\!\left(-10\cdot\operatorname{clamp}\big(g(X_{\mathrm{sar}}),0,0.3\big)\right),
 \label{eq:flatness}
\end{equation}
where the clamp keeps isolated speckle outliers from counting as evidence. The hallucination-frequency index is the share of a candidate's gradient energy that falls in SAR-flat regions, and the gradient-magnitude correlation compares speckle-suppressed gradient maps:
\begin{align}
 \operatorname{HFI}(Y)&=\operatorname{clip}\!\left(\frac{\sum M_{\mathrm{flat}}\odot g(Y)}{\sum g(Y)+\epsilon},0,1\right),\label{eq:hfi}\\
 \operatorname{GMC}(Y)&=\operatorname{Pearson}\!\left(\tilde g(Y),\tilde g(X_{\mathrm{sar}})\right).\label{eq:gmc}
\end{align}
GMC is invariant to affine contrast changes and therefore tolerant to the modality gap. The candidates are ranked by
\begin{equation}
 S(Y)=\operatorname{GMC}(Y)-w\cdot\operatorname{HFI}(Y),\quad k^*=\arg\max_k S(Y_k),
 \label{eq:toposcore}
\end{equation}
with $w=5$. The score can be read as a heuristic MAP selection among the samples. After log compression, speckle becomes additive with a constant offset that cancels in spatial derivatives, and pooling reduces its variance, so $\tilde g(X_{\mathrm{sar}})$ is a noisy cue to edge location and GMC acts as a likelihood term for edge co-location; HFI penalizes gradient energy without SAR support and acts as a prior against invented structure. This reading motivates~(\ref{eq:toposcore}) but is not a guarantee (see Appendix~\ref{app:theory}).

\subsection{SAR-Guided Frequency Gate}
Selection cannot remove unsupported detail inside the chosen candidate $Y_{\mathrm{best}}=Y_{k^*}$. We therefore split it with a $5\times5$ Gaussian $G$ (binomial kernel, $\sigma\approx1$~px) and keep the residual only where the SAR shows edges:
\begin{equation}
\begin{aligned}
 Y_{\mathrm{low}}&=G\ast Y_{\mathrm{best}},\quad M_{\mathrm{edge}}=\operatorname{sigm}\!\big(s\,(g(X_{\mathrm{sar}})-t)\big),\\
 Y_{\mathrm{final}}&=Y_{\mathrm{low}}+M_{\mathrm{edge}}\odot\left(Y_{\mathrm{best}}-Y_{\mathrm{low}}\right),
\end{aligned}
\label{eq:gate}
\end{equation}
with the logistic function $\operatorname{sigm}$, $s=5$, and $t=0.1$. $Y_{\mathrm{low}}$ is always kept, so large-scale appearance is unchanged.

\paragraph{Hyperparameters.} $K=5$, $w=5$, and $r=3$ were chosen at the knee of one-factor-at-a-time sweeps on a tuning subset before this study, and $s$ and $t$ were set heuristically. All five were then frozen: no value was changed for any backbone or data set in this work.

\section{Experiments}
\subsection{Protocol}
We report PSNR, SSIM \cite{wang2004image}, and pixel-level precision, recall, and F1 of Canny edge maps \cite{canny1986computational} (Otsu-based thresholds) against the optical reference (definitions in Appendix~\ref{app:metrics}). Edge precision falls when an output contains edges absent from the reference, so it directly measures added structure. Each image's metric is first averaged over the three seeds; differences are tested with paired Wilcoxon signed-rank tests and 10,000-sample bootstrap 95\% confidence intervals (CI), and Table~\ref{tab:main} gives the spread across seeds. COP-GEN is evaluated on 200 patches per set with three sampling seeds; its outputs and references are compared after a per-image 2--98\% stretch because its radiometry differs from SEN1-2. Code, frozen split lists, checkpoints, and per-image results are released at \url{https://github.com/researchworksofsomeone/SAR-TopoGuard_Repo}.

\begin{table}[t]
\caption{Pix2Pix results (mean $\pm$ std over three training seeds). Best value per set and metric in bold.}
\label{tab:main}
\centering
\small
\setlength{\tabcolsep}{6pt}
\begin{tabular}{@{}llccccc@{}}
\toprule
Set & Method & PSNR (dB) & SSIM & Edge P & Edge R & Edge F1\\
\midrule
Test & B1 & $11.36\pm.13$ & $.0792\pm.005$ & $.267\pm.002$ & $.304\pm.003$ & $.279\pm.000$\\
(500 patches) & B2 & $11.40\pm.05$ & $.0803\pm.001$ & $.268\pm.001$ & $\mathbf{.306}\pm.001$ & $\mathbf{.281}\pm.001$\\
 & Ours & $\mathbf{11.42}\pm.12$ & $\mathbf{.0826}\pm.005$ & $\mathbf{.268}\pm.002$ & $.294\pm.003$ & $.275\pm.000$\\
\midrule
Summer & B1 & $10.43\pm.27$ & $.0813\pm.005$ & $.146\pm.002$ & $\mathbf{.299}\pm.010$ & $\mathbf{.193}\pm.001$\\
(500 patches) & B2 & $10.23\pm.20$ & $\mathbf{.0877}\pm.006$ & $\mathbf{.151}\pm.004$ & $.282\pm.016$ & $\mathbf{.193}\pm.001$\\
 & Ours & $\mathbf{10.44}\pm.28$ & $.0870\pm.005$ & $.148\pm.002$ & $.289\pm.010$ & $\mathbf{.193}\pm.001$\\
\midrule
Winter & B1 & $13.10\pm.26$ & $.1505\pm.005$ & $.129\pm.001$ & $.309\pm.003$ & $.169\pm.001$\\
(200 pairs) & B2 & $13.09\pm.19$ & $.1530\pm.009$ & $.129\pm.001$ & $\mathbf{.311}\pm.002$ & $\mathbf{.170}\pm.001$\\
 & Ours & $\mathbf{13.17}\pm.26$ & $\mathbf{.1640}\pm.005$ & $\mathbf{.130}\pm.001$ & $.297\pm.003$ & $.168\pm.001$\\
\bottomrule
\end{tabular}
\end{table}

\begin{table}[t]
\caption{Ablation: $\Delta$PSNR (dB) / $\Delta$SSIM ($\times10^{3}$) vs.\ the plain output. Random: random candidate; Verifier: selection without gate; Gate: random candidate + gate; $|\Delta|$ cand.: mean per-image PSNR difference (dB) between two candidates.}
\label{tab:ablation}
\centering
\small
\setlength{\tabcolsep}{7pt}
\begin{tabular}{@{}lccccc@{}}
\toprule
 & \multicolumn{3}{c}{Pix2Pix} & \multicolumn{2}{c}{COP-GEN, winter set}\\
\cmidrule(lr){2-4}\cmidrule(lr){5-6}
Configuration & Test & Summer & Winter & Base & Beta\\
\midrule
Random & $+.003/+0.0$ & $-.007/+0.0$ & $-.002/-0.1$ & $-.04/-1.6$ & $+.01/+1.9$\\
Verifier & $+.001/+0.2$ & $-.032/+0.1$ & $-.017/-0.1$ & $\mathbf{+.26}/+6.4$ & $+.07/+8.6$\\
Gate & $\mathbf{+.062}/+3.2$ & $\mathbf{+.038}/+5.7$ & $\mathbf{+.093}/\mathbf{+13.5}$ & $+.02/+3.8$ & $+.10/+9.8$\\
Full & $+.060/\mathbf{+3.4}$ & $+.012/\mathbf{+5.7}$ & $+.077/\mathbf{+13.5}$ & $\mathbf{+.33}/\mathbf{+12.3}$ & $\mathbf{+.16}/\mathbf{+16.7}$\\
\midrule
$|\Delta|$ cand. & 0.06 & 0.07 & 0.08 & 1.19 & 1.32\\
\bottomrule
\end{tabular}
\end{table}

\begin{table}[t]
\caption{Other backbones: plain PSNR (dB) and gain from \method\ with bootstrap 95\% CI; $\Delta$SSIM in $10^{-3}$; \% = share of images with higher PSNR.}
\label{tab:transfer}
\centering
\resizebox{\linewidth}{!}{%
\begin{tabular}{@{}lcccc@{\hspace{18pt}}lcccc@{}}
\toprule
\multicolumn{5}{c}{CycleGAN} & \multicolumn{5}{c}{CUT}\\
\cmidrule(r{10pt}){1-5}\cmidrule(l){6-10}
Set & Plain & $\Delta$PSNR [95\% CI] & $\Delta$SSIM & \% & Set & Plain & $\Delta$PSNR [95\% CI] & $\Delta$SSIM & \%\\
\midrule
Test & 9.55 & $+.065$ [$+.057$, $+.074$] & $-0.0$ & 78 & Test & 9.95 & $+.053$ [$+.049$, $+.057$] & $+1.9$ & 92\\
Summer & 8.59 & $+.013$ [$+.004$, $+.022$] & $-2.0$ & 58 & Summer & 9.06 & $+.025$ [$+.022$, $+.027$] & $+2.4$ & 86\\
Winter & 11.36 & $+.051$ [$+.030$, $+.072$] & $+4.2$ & 68 & Winter & 11.33 & $+.076$ [$+.064$, $+.088$] & $+9.3$ & 91\\
\midrule
\multicolumn{5}{c}{COP-GEN base} & \multicolumn{5}{c}{COP-GEN Beta}\\
\cmidrule(r{10pt}){1-5}\cmidrule(l){6-10}
Test$^\dagger$ & 8.54 & $+.072$ [$-.038$, $+.182$] & $+1.2$ & 51 & Test$^\dagger$ & 8.96 & $-.042$ [$-.123$, $+.037$] & $+1.6$ & 44\\
Summer$^\dagger$ & 9.02 & $+.072$ [$-.040$, $+.186$] & $+6.6$ & 54 & Summer$^\dagger$ & 8.80 & $+.087$ [$+.002$, $+.171$] & $+16.9$ & 58\\
Winter & 8.91 & $+.330$ [$+.186$, $+.479$] & $+12.3$ & 68 & Winter & 9.64 & $+.164$ [$-.012$, $+.337$] & $+16.7$ & 57\\
\bottomrule
\multicolumn{10}{@{}l@{}}{$^\dagger$VV only; VV is also given as the VH input of COP-GEN.}
\end{tabular}}
\end{table}

\subsection{Results}
\paragraph{Pix2Pix.} Table~\ref{tab:main} shows that \method\ improves PSNR and SSIM on all three sets and for every seed. The paired gains are $+0.060$~dB [$+0.055$, $+0.065$] and $+0.0034$ SSIM on the test set (87\% and 98\% of images improve), $+0.012$~dB and $+0.0057$ on the summer scene, and $+0.077$~dB [$+0.065$, $+0.089$] and $+0.0135$ on the winter set, where SSIM improves for all 200 images (all $p<10^{-6}$; full paired tests in Table~\ref{tab:paired}). Edge precision rises on every set ($+0.0013$ to $+0.0019$, $p<10^{-25}$) while edge recall falls by about 0.01, so edge F1 drops by $0.0037$ ($-1.3\%$) on the test set and by $0.0006$ or less on both OOD sets. The layer thus removes edges that are absent from the reference, together with some faint true edges, which is the intended trade-off. The edge-loss baseline B2 behaves differently: it raises edge F1 slightly ($+0.0015$ on the test set) but gives no significant PSNR gain on the test or winter sets and loses $0.19$~dB on the summer scene; post-hoc gating and training-time edge supervision are therefore complementary rather than interchangeable.

\paragraph{Ablation.} Table~\ref{tab:ablation} isolates the components. For Pix2Pix, a random MC-dropout candidate is statistically indistinguishable from the deterministic output, and verifier selection alone changes PSNR by at most 0.03~dB, and slightly lowers it on both OOD sets; the gate alone reproduces the full gain. The last row explains why: the mean per-image PSNR difference between two candidates is only 0.06--0.08~dB for MC dropout but 1.2--1.3~dB for COP-GEN. With diverse diffusion samples, selection becomes useful: on the winter set the verifier alone adds $+0.26$~dB to COP-GEN base, and verifier plus gate gives the largest gains for both checkpoints. Consistent with this, the verifier weight hardly matters for Pix2Pix: over $w\in[0,20]$, PSNR varies by less than 0.002~dB (Fig.~\ref{fig:wsweep}).

\paragraph{Other backbones.} With the same frozen settings, \method\ improves CycleGAN and CUT on all sets (Table~\ref{tab:transfer}, $p<10^{-5}$); SSIM falls slightly only for CycleGAN on the summer scene. For COP-GEN, the largest gains occur on the winter set, whose SAR has true VV and VH: $+0.33$~dB for base ($p<10^{-6}$) and $+0.16$~dB for Beta (Wilcoxon $p=0.026$, bootstrap CI touching zero), with SSIM gains for both ($p<10^{-9}$). On the VV-only SEN1-2 sets, the VH input is missing and the PSNR changes are not significant ($p\geq0.057$), although SSIM still improves on the summer scene for both checkpoints. A frozen DINOv2 probe \cite{oquab2024dinov2} shows no significant change for any COP-GEN setting ($|\Delta|<0.01$, $p>0.3$; Table~\ref{tab:copgen}), so we do not claim a semantic gain.

\paragraph{Cost.} The layer has no trainable parameters and needs 0.05~GFLOPs per patch. On an NVIDIA RTX PRO 6000, five Pix2Pix passes (54.4~M parameters, 12.1~GFLOPs each) take 20.0~ms, and the verifier and gate add 0.33~ms and 0.07~ms. CycleGAN and CUT use 11.4~M parameters and 99.1~GFLOPs per pass. Training one seed takes about 9~min for Pix2Pix and 30--45~min for CUT and CycleGAN.

\begin{figure}[t]
\centering
\includegraphics[width=\linewidth]{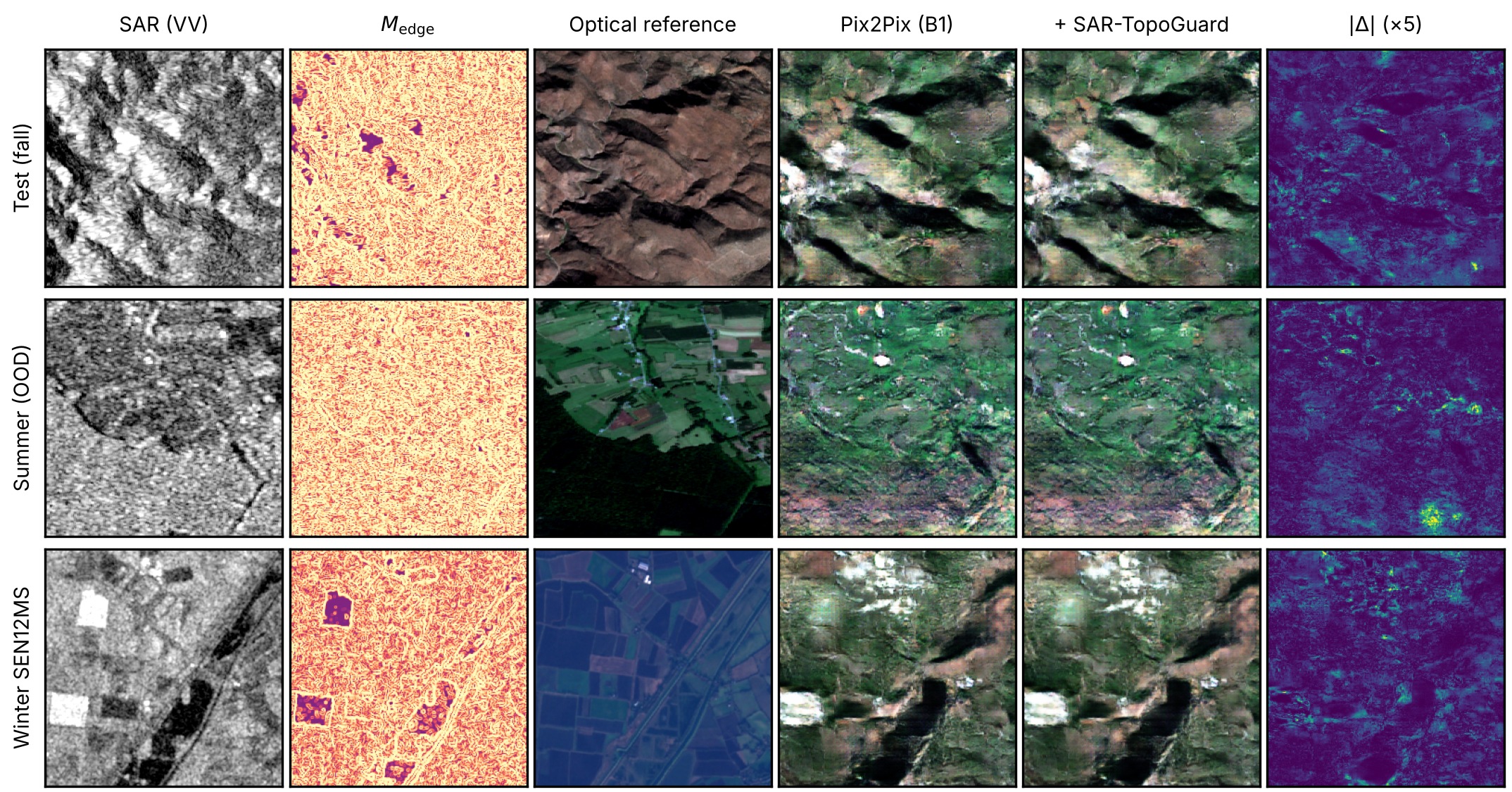}
\caption{Examples from the test set, the summer scene, and the winter set (seed 0; for each set, the patch with the largest change). $M_{\mathrm{edge}}$ is the SAR edge mask of (\ref{eq:gate}) (bright: residual kept), and $|\Delta|$ is the absolute difference between the Pix2Pix output and \method, amplified five times. Changes concentrate in SAR-flat areas (dark in $M_{\mathrm{edge}}$), such as fields and smooth slopes, where the gate removes high-frequency texture; the large-scale appearance is unchanged.}
\label{fig:qual}
\end{figure}

\subsection{Qualitative Assessment}
Fig.~\ref{fig:qual} shows that \method\ does not repaint the scene: the low-frequency content of the generator is kept, and differences appear where the SAR is flat, for example in homogeneous fields and smooth slopes, where the generator adds speckle-like texture. Strong SAR edges, such as field boundaries and roads, keep their detail. Because speckle makes $g(X_{\mathrm{sar}})$ exceed $t$ in many textured areas, the gate acts mainly on large flat regions, which also explains why the gains are modest.

\section{Conclusion}
\method\ is a training-free layer that selects and gates the outputs of frozen SAR-to-optical generators against the SAR measurement. Across three GANs, two diffusion checkpoints, a scene-disjoint test set, a held-out season, and an independent winter set, it gives small but consistent PSNR and SSIM gains and higher edge precision, at the cost of lower edge recall. The ablation shows when each component helps: SAR-guided gating is effective even when the sampler gives almost identical candidates, and SAR-conditioned selection pays off when samples differ.

\paragraph{Limitations.} 1) The gains are small (below 0.1~dB for GANs), and the backbones have low absolute fidelity (9--13~dB PSNR), so the results show relative improvements, not state-of-the-art translation. 2) The five hyperparameters are hand-set ($s$ and $t$ heuristically) and were fixed on an earlier tuning subset of the same source whose patch list was not kept, so overlap with the summer scene cannot be excluded; the winter set is fully independent. 3) Inference costs $K$ generator passes, and the sampler differs per backbone. 4) The evaluation covers Sentinel-1/2 at 10~m with VV-only training data, and the summer set is a single scene; other sensors, resolutions, and polarizations are untested. 5) SIFT-based matching rates are near zero for all methods on these data, so we could not test registration utility directly.

Future work will test the layer on a second sensor pair (e.g., Gaofen-3 or TerraSAR-X with optical references), replace the fixed gate threshold with one set from the local speckle statistics, so that the gate also acts in textured areas, and measure the effect on registration and change-detection pipelines \cite{zitova2003image}.

\section*{Code and Data Availability}
Code, frozen split lists, checkpoints, and per-image metrics: \url{https://github.com/researchworksofsomeone/SAR-TopoGuard_Repo}. SEN1-2 \cite{schmitt2018sen12} and SEN12MS \cite{schmitt2019sen12ms} are publicly available; COP-GEN checkpoints are used as released \cite{espinosa2026copgen,espinosa2025copgenbeta}.

\section*{Conflict of Interest}
The authors declare that they have no conflicts of interest.

\appendix
\renewcommand{\thetable}{A\arabic{table}}
\renewcommand{\thefigure}{A\arabic{figure}}
\renewcommand{\theequation}{A\arabic{equation}}
\setcounter{table}{0}\setcounter{figure}{0}\setcounter{equation}{0}

\section{Reproducibility}\label{app:repro}
\paragraph{Data.} The SEN1-2 subset \cite{schmitt2018sen12} holds 16,000 terrain-stratified $256\times256$ pairs (8-bit PNG; Sentinel-1 VV in dB clipped to $[-25,0]$; Sentinel-2 B4/B3/B2) from 29 fall scenes and one summer scene. Scene identifiers are parsed from the file names, and whole scenes are assigned to splits (Table~\ref{tab:split}); a pair-hash check confirms that no patch occurs in two splits. The winter set consists of 200 SEN12MS \cite{schmitt2019sen12ms} pairs from 31 scenes (GeoTIFF; VV and VH in dB; 13-band Sentinel-2, of which B4/B3/B2 are used). The split lists are frozen and released.

\paragraph{Preprocessing.} SEN1-2 stores dB values, so the 8-bit values are mapped back to $[-25,0]$~dB and divided by 10 to obtain $\log_{10}$ intensity; no second logarithm is taken. $P_2$ and $P_{98}$ of each modality are estimated on the training set (4,096 random pixels per patch), and (\ref{eq:normalization}) maps both modalities to $[0,1]$. For the winter set, whose optical bands are 16-bit reflectances, the percentiles are estimated on that set (optical $P_2/P_{98}=412/2508$). The GANs were trained on VV only, so VV fills both SAR input channels for every set.

\begin{figure}[t]
\centering
\includegraphics[width=0.85\linewidth]{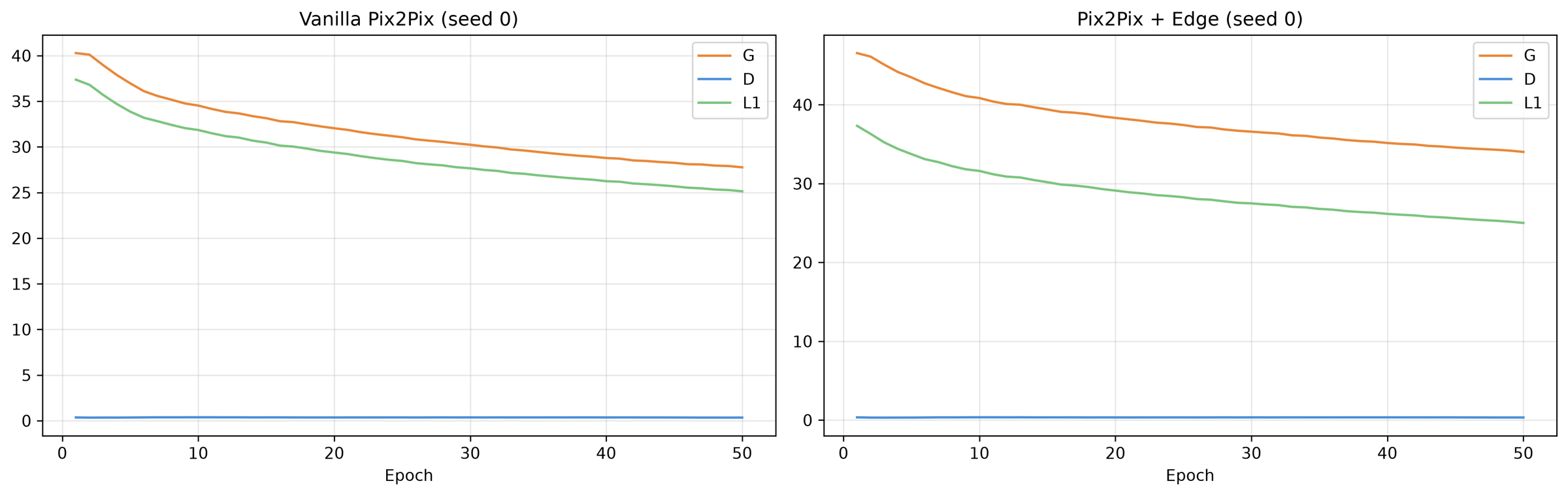}
\caption{Training losses of Pix2Pix (B1, left) and Pix2Pix with edge loss (B2, right), seed 0. G: generator; D: discriminator; L1: reconstruction term.}
\label{fig:train}
\end{figure}

\paragraph{Training.} Pix2Pix (B1; $\lambda_{L1}=100$, real-label smoothing 0.9) and the edge-loss variant (B2; additionally $\lambda_e=10$) are trained for 50 epochs with batch 8, Adam ($2\times10^{-4}$, $\beta=(0.5,0.999)$). CycleGAN \cite{zhu2017unpaired} ($\lambda_{\mathrm{cyc}}=10$, $\lambda_{\mathrm{idt}}=0.5$) and CUT \cite{park2020contrastive} (PatchNCE weight 1, $\tau=0.07$, 256 patches, five layers) use LSGAN losses, batch 8, Adam ($2\times10^{-4}$), 30 epochs with linear decay after epoch 15, and dropout ($p=0.5$) in the residual blocks so that MC dropout can sample them. Every model is trained with seeds 0, 1, and 2; the last-epoch checkpoint is evaluated, and no test data are used for model selection. Fig.~\ref{fig:train} shows the losses of B1 and B2 (seed 0): generator and $L_1$ losses decrease monotonically over 50 epochs, and the discriminator loss stays stable, with no sign of collapse; the other seeds behave the same way.

\paragraph{COP-GEN.} The released base and Beta checkpoints \cite{espinosa2026copgen,espinosa2025copgenbeta} generate all non-Sentinel-1 modalities jointly from Sentinel-1 (50 DPM-Solver++ steps). The $K=5$ candidates are five noise seeds for the same input; the plain output is candidate~0. On the SEN1-2 sets, VV is also passed as VH. 200 patches per set and three sampling seeds are used. Outputs and references are compared after a per-image 2--98\% stretch.

\paragraph{Environment.} One NVIDIA RTX PRO 6000 Blackwell GPU (PyTorch, CUDA). The released code implements exactly the equations printed in this paper (normalization, channel-mean Sobel gradient, pooled Pearson GMC, normalized HFI, Gaussian gate, Otsu-based Canny), and every experiment was run from scratch on the scene-disjoint split.

\section{Metrics}\label{app:metrics}
PSNR and SSIM \cite{wang2004image} (7$\times$7 window, data range 1) are computed on RGB in $[0,1]$. Canny maps \cite{canny1986computational} of the gray images use thresholds $(0.67T,1.33T)$, where $T$ is the Otsu threshold. With generated edges $E_g$ and reference edges $E_r$,
\begin{equation}
P=\frac{|E_g\cap E_r|}{|E_g|},\quad R=\frac{|E_g\cap E_r|}{|E_r|},\quad F_1=\frac{2PR}{P+R}.
\end{equation}
Edge-IoU uses the same maps after one-pixel dilation $D$: $|D(E_g)\cap D(E_r)|/|D(E_g)\cup D(E_r)|$. The structural matching rate (SMR) is the number of mutual nearest-neighbor SIFT matches \cite{lowe2004distinctive} that pass Lowe's ratio test (0.75), divided by the smaller keypoint count. The DINOv2 probe \cite{oquab2024dinov2} is the cosine similarity of frozen DINOv2-base pooled embeddings (224~px input) of output and reference. For every metric, a paired difference is computed per image after averaging over seeds; we report its mean, a 10,000-sample bootstrap 95\% CI, and the Wilcoxon signed-rank $p$.

\section{Analysis of the Verifier and Gate}\label{app:theory}
\paragraph{Pooling.} After log compression, write the SAR gradient response as $\hat g(p)=u(p)+\xi(p)$, with structural part $u$ ($L$-Lipschitz) and speckle fluctuation $\xi$ of zero mean and variance at most $\sigma_\xi^2$. If the fluctuations in an $r\times r$ window are uncorrelated, the pooled response $\tilde g_r(p)$ satisfies
\begin{equation}
\mathbb E\big[(\tilde g_r(p)-u(p))^2\big]\le \frac{\sigma_\xi^2}{r^2}+L^2d_r^2,
\end{equation}
where $d_r$ is the mean distance from $p$ within the window. Pooling thus trades speckle variance for localization bias; $r=3$ is the smallest window that reduces the variance ninefold. Sobel responses of neighboring pixels are correlated, so the reduction is smaller in practice, and the bound only motivates the choice.

\paragraph{Gate.} With $B(Y)=G\ast Y$, the gate $\mathcal T(Y)=(1-M)\odot B(Y)+M\odot Y$ is a pointwise convex combination of the candidate and its low-pass version. It never adds structure: $\mathcal T(Y)=Y$ where $M\to1$, $\mathcal T(Y)=B(Y)$ where $M\to0$, and $|\partial M/\partial g|\le s/4$, so the mask changes smoothly with the SAR gradient. Removing residual in SAR-flat regions raises PSNR wherever that residual is absent from the reference and lowers it where the reference does contain such detail; the observed PSNR and SSIM gains together with the lower edge recall indicate that the first case dominates.

\section{Full Results}\label{app:results}
\begin{figure}[t]
\centering
\includegraphics[width=0.85\linewidth]{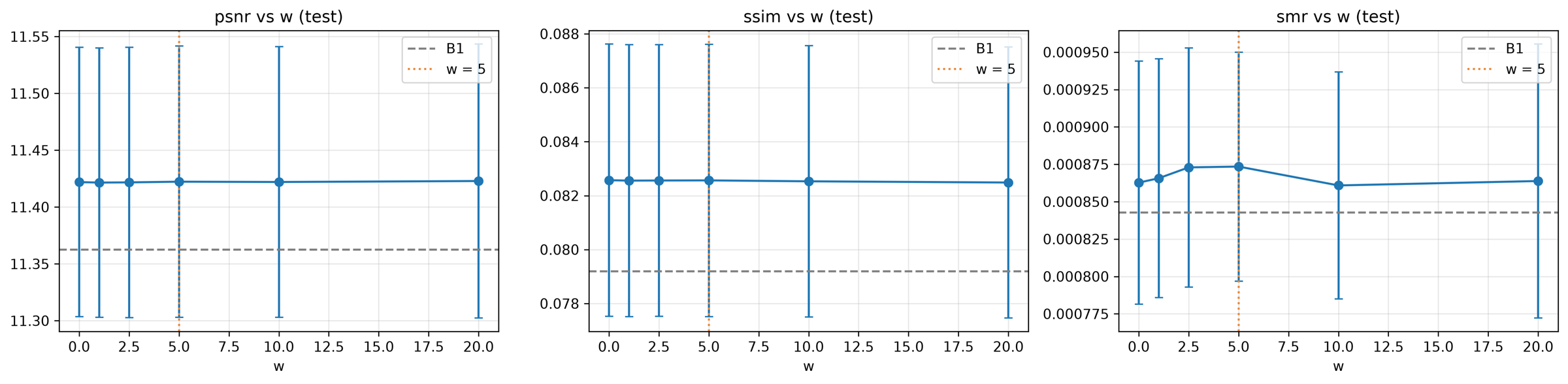}
\caption{Pix2Pix test-set PSNR, SSIM, and SMR versus the verifier weight $w$ (mean and spread over three seeds; dashed: B1; dotted: $w=5$). All other settings are fixed.}
\label{fig:wsweep}
\end{figure}

Table~\ref{tab:p2p} extends Table~\ref{tab:main} with the tuning set, Edge-IoU, and SMR. SMR is below 0.003 for all methods because few SIFT keypoints of the translated images match the optical reference, so it cannot separate the methods (the Edge-IoU pattern follows edge F1). Table~\ref{tab:paired} gives the paired tests against B1. Table~\ref{tab:copgen} gives the COP-GEN results with all ablations and the DINOv2 probe. Fig.~\ref{fig:wsweep} shows that the verifier weight $w$ has almost no effect on Pix2Pix, consistent with the ablation in Table~\ref{tab:ablation}: MC-dropout candidates are nearly identical, so the choice among them hardly matters.

\begin{table}[t]
\caption{Pix2Pix results per set (mean $\pm$ std over three training seeds); SMR and mean number of SIFT matches are given without std.}
\label{tab:p2p}
\centering
\resizebox{\linewidth}{!}{%
\begin{tabular}{@{}llcccccccc@{}}
\toprule
Set & Method & PSNR (dB) & SSIM & Edge P & Edge R & Edge F1 & Edge-IoU & SMR ($\times10^{3}$) & Matches\\
\midrule
Tune & B1 & $10.56\pm0.18$ & $.0700\pm.0009$ & $.270\pm.002$ & $.294\pm.007$ & $.278\pm.002$ & $.639\pm.007$ & $0.92$ & $0.90$\\
 & B2 & $10.71\pm0.14$ & $.0717\pm.0024$ & $.271\pm.003$ & $.296\pm.001$ & $.280\pm.001$ & $.646\pm.002$ & $0.86$ & $0.82$\\
 & Ours & $10.63\pm0.18$ & $.0737\pm.0011$ & $.272\pm.002$ & $.280\pm.007$ & $.273\pm.002$ & $.623\pm.007$ & $0.94$ & $0.93$\\
\midrule
Test & B1 & $11.36\pm0.13$ & $.0792\pm.0050$ & $.267\pm.002$ & $.304\pm.003$ & $.279\pm.000$ & $.649\pm.002$ & $0.84$ & $0.77$\\
 & B2 & $11.40\pm0.05$ & $.0803\pm.0012$ & $.268\pm.001$ & $.306\pm.001$ & $.281\pm.001$ & $.651\pm.001$ & $0.85$ & $0.76$\\
 & Ours & $11.42\pm0.12$ & $.0826\pm.0050$ & $.268\pm.002$ & $.294\pm.003$ & $.275\pm.000$ & $.637\pm.003$ & $0.87$ & $0.80$\\
\midrule
Summer & B1 & $10.43\pm0.27$ & $.0813\pm.0051$ & $.146\pm.002$ & $.299\pm.010$ & $.193\pm.001$ & $.391\pm.002$ & $1.76$ & $0.83$\\
 & B2 & $10.23\pm0.20$ & $.0877\pm.0055$ & $.151\pm.004$ & $.282\pm.016$ & $.193\pm.001$ & $.391\pm.003$ & $2.02$ & $0.92$\\
 & Ours & $10.44\pm0.28$ & $.0870\pm.0053$ & $.148\pm.002$ & $.289\pm.010$ & $.193\pm.001$ & $.390\pm.002$ & $2.00$ & $0.93$\\
\midrule
Winter & B1 & $13.10\pm0.26$ & $.1505\pm.0046$ & $.129\pm.001$ & $.309\pm.003$ & $.169\pm.001$ & $.342\pm.002$ & $2.31$ & $0.48$\\
 & B2 & $13.09\pm0.19$ & $.1530\pm.0087$ & $.129\pm.001$ & $.311\pm.002$ & $.170\pm.001$ & $.345\pm.002$ & $2.33$ & $0.45$\\
 & Ours & $13.17\pm0.26$ & $.1640\pm.0049$ & $.130\pm.001$ & $.297\pm.003$ & $.168\pm.001$ & $.339\pm.003$ & $2.06$ & $0.52$\\
\bottomrule
\end{tabular}}
\end{table}

\begin{table}[t]
\caption{Paired differences against B1 (per image, seed-averaged). $\Delta$PSNR with bootstrap 95\% CI and Wilcoxon $p$; other columns in $10^{-3}$ with Wilcoxon $p$.}
\label{tab:paired}
\centering
\resizebox{\linewidth}{!}{%
\begin{tabular}{@{}llcccccc@{}}
\toprule
Set & Method & $\Delta$PSNR (dB) [95\% CI] & $p$ & $\Delta$SSIM & $\Delta$Edge P & $\Delta$Edge R & $\Delta$Edge F1\\
\midrule
Tune & Ours & $+0.074$ [$+0.067$, $+0.081$] & $<10^{-35}$ & $+3.7$ ($<10^{-42}$) & $+1.6$ ($<10^{-28}$) & $-13.7$ ($<10^{-42}$) & $-5.3$ ($<10^{-40}$)\\
 & B2 & $+0.148$ [$+0.090$, $+0.208$] & $<10^{-5}$ & $+1.6$ ($<10^{-3}$) & $+0.7$ ($0.02$) & $+1.7$ ($<10^{-3}$) & $+2.1$ ($<10^{-8}$)\\
Test & Ours & $+0.060$ [$+0.055$, $+0.065$] & $<10^{-62}$ & $+3.4$ ($<10^{-82}$) & $+1.3$ ($<10^{-49}$) & $-9.8$ ($<10^{-82}$) & $-3.7$ ($<10^{-79}$)\\
 & B2 & $+0.035$ [$-0.005$, $+0.076$] & $0.96$ & $+1.2$ ($<10^{-3}$) & $+0.8$ ($<10^{-5}$) & $+2.3$ ($<10^{-7}$) & $+1.5$ ($<10^{-12}$)\\
Summer & Ours & $+0.012$ [$+0.007$, $+0.017$] & $<10^{-6}$ & $+5.7$ ($<10^{-82}$) & $+1.9$ ($<10^{-72}$) & $-10.1$ ($<10^{-82}$) & $-0.5$ ($<10^{-5}$)\\
 & B2 & $-0.194$ [$-0.240$, $-0.147$] & $<10^{-20}$ & $+6.4$ ($<10^{-52}$) & $+4.4$ ($<10^{-62}$) & $-16.5$ ($<10^{-69}$) & $-0.0$ ($0.76$)\\
Winter & Ours & $+0.077$ [$+0.065$, $+0.089$] & $<10^{-22}$ & $+13.5$ ($<10^{-33}$) & $+1.7$ ($<10^{-25}$) & $-11.7$ ($<10^{-27}$) & $-0.6$ ($<10^{-4}$)\\
 & B2 & $-0.004$ [$-0.104$, $+0.090$] & $0.20$ & $+2.5$ ($0.007$) & $+0.9$ ($0.005$) & $+1.7$ ($<10^{-3}$) & $+1.8$ ($<10^{-7}$)\\
\bottomrule
\end{tabular}}
\end{table}

\begin{table}[t]
\caption{COP-GEN: plain PSNR, $\Delta$PSNR (dB) of each configuration vs.\ the plain sample, and $\Delta$SSIM, $\Delta$Edge F1, and $\Delta$DINOv2 cosine ($10^{-3}$) of the full method, with Wilcoxon $p$ (200 patches, three sampling seeds).}
\label{tab:copgen}
\centering
\resizebox{\linewidth}{!}{%
\begin{tabular}{@{}llcccccccc@{}}
\toprule
Model & Set & Plain & Random & Verifier & Gate & Full ($p$) & $\Delta$SSIM & $\Delta$Edge F1 & $\Delta$DINOv2\\
\midrule
Base & Test$^\dagger$ & 8.54 & $-0.05$ & $+0.04$ & $-0.02$ & $\mathbf{+0.07}$ ($0.35$) & $+1.2$ ($0.35$) & $+4.1$ ($0.06$) & $-0.5$ ($0.81$)\\
 & Summer$^\dagger$ & 9.02 & $-0.03$ & $+0.04$ & $+0.01$ & $\mathbf{+0.07}$ ($0.25$) & $+6.6$ ($0.001$) & $+5.2$ ($<10^{-4}$) & $+7.4$ ($0.34$)\\
 & Winter & 8.91 & $-0.04$ & $+0.26$ & $+0.02$ & $\mathbf{+0.33}$ ($<10^{-6}$) & $+12.3$ ($<10^{-9}$) & $-0.3$ ($0.93$) & $-6.8$ ($0.35$)\\
Beta & Test$^\dagger$ & 8.96 & $-0.02$ & $-0.09$ & $+0.03$ & $\mathbf{-0.04}$ ($0.07$) & $+1.6$ ($0.11$) & $-1.6$ ($<10^{-5}$) & $+2.4$ ($0.55$)\\
 & Summer$^\dagger$ & 8.80 & $-0.02$ & $+0.05$ & $+0.03$ & $\mathbf{+0.09}$ ($0.06$) & $+16.9$ ($<10^{-22}$) & $+2.1$ ($<10^{-3}$) & $-0.6$ ($0.75$)\\
 & Winter & 9.64 & $+0.01$ & $+0.07$ & $+0.10$ & $\mathbf{+0.16}$ ($0.03$) & $+16.7$ ($<10^{-15}$) & $-3.3$ ($0.003$) & $+6.0$ ($0.65$)\\
\bottomrule
\multicolumn{10}{@{}l@{}}{$^\dagger$SEN1-2 sets are VV only; VV is also given as the VH input.}
\end{tabular}}
\end{table}

\bibliographystyle{unsrt}
\bibliography{references}

\begin{thebibliography}{10}

\bibitem{moreira2013tutorial}
Alberto Moreira, Pau Prats-Iraola, Marwan Younis, Gerhard Krieger, Irena
  Hajnsek, and Konstantinos~P. Papathanassiou.
\newblock A tutorial on synthetic aperture radar.
\newblock {\em IEEE Geoscience and Remote Sensing Magazine}, 1(1):6--43, 2013.
\newblock doi:
  \href{https://doi.org/10.1109/MGRS.2013.2248301}{\nolinkurl{10.1109/MGRS.2013.2248301}}.

\bibitem{schmitt2018sen12}
Michael Schmitt, Lloyd~H. Hughes, Chunping Qiu, and Xiao~Xiang Zhu.
\newblock {SEN1-2}: A weakly supervised pair of {Sentinel-1} and {Sentinel-2}
  datasets.
\newblock {\em ISPRS Annals of Photogrammetry, Remote Sensing and Spatial
  Information Sciences}, IV-3:193--200, 2018.
\newblock doi:
  \href{https://doi.org/10.5194/isprs-annals-IV-3-2018-193}{\nolinkurl{10.5194/isprs-annals-IV-3-2018-193}}.

\bibitem{isola2017image}
Phillip Isola, Jun-Yan Zhu, Tinghui Zhou, and Alexei~A. Efros.
\newblock Image-to-image translation with conditional adversarial networks.
\newblock In {\em Proceedings of the IEEE Conference on Computer Vision and
  Pattern Recognition (CVPR)}, pages 1125--1134, 2017.
\newblock doi:
  \href{https://doi.org/10.1109/CVPR.2017.632}{\nolinkurl{10.1109/CVPR.2017.632}}.

\bibitem{zitova2003image}
Barbara Zitova and Jan Flusser.
\newblock Image registration methods: A survey.
\newblock {\em Image and Vision Computing}, 21(11):977--1000, 2003.
\newblock doi:
  \href{https://doi.org/10.1016/S0262-8856(03)00137-9}{\nolinkurl{10.1016/S0262-8856(03)00137-9}}.

\bibitem{wang2004image}
Zhou Wang, Alan~C. Bovik, Hamid~R. Sheikh, and Eero~P. Simoncelli.
\newblock Image quality assessment: From error visibility to structural
  similarity.
\newblock {\em IEEE Transactions on Image Processing}, 13(4):600--612, 2004.
\newblock doi:
  \href{https://doi.org/10.1109/TIP.2003.819861}{\nolinkurl{10.1109/TIP.2003.819861}}.

\bibitem{gal2016dropout}
Yarin Gal and Zoubin Ghahramani.
\newblock Dropout as a {Bayesian} approximation: Representing model uncertainty
  in deep learning.
\newblock In {\em Proceedings of the 33rd International Conference on Machine
  Learning (ICML)}, pages 1050--1059, 2016.
\newblock doi:
  \href{https://doi.org/10.48550/arXiv.1506.02142}{\nolinkurl{10.48550/arXiv.1506.02142}}.

\bibitem{zhu2017unpaired}
Jun-Yan Zhu, Taesung Park, Phillip Isola, and Alexei~A. Efros.
\newblock Unpaired image-to-image translation using cycle-consistent
  adversarial networks.
\newblock In {\em Proceedings of the IEEE International Conference on Computer
  Vision (ICCV)}, pages 2223--2232, 2017.
\newblock doi:
  \href{https://doi.org/10.1109/ICCV.2017.244}{\nolinkurl{10.1109/ICCV.2017.244}}.

\bibitem{park2020contrastive}
Taesung Park, Alexei~A. Efros, Richard Zhang, and Jun-Yan Zhu.
\newblock Contrastive learning for unpaired image-to-image translation.
\newblock In {\em European Conference on Computer Vision (ECCV)}, pages
  319--345, 2020.
\newblock doi:
  \href{https://doi.org/10.1007/978-3-030-58545-7_19}{\nolinkurl{10.1007/978-3-030-58545-7_19}}.

\bibitem{zhang2025segcyclegan}
Hannuo Zhang, Huihui Li, Jiarui Lin, Yujie Zhang, Jianghua Fan, Hang Liu, and
  Kun Liu.
\newblock {Seg-CycleGAN}: {SAR}-to-optical image translation guided by a
  downstream task.
\newblock {\em IEEE Geoscience and Remote Sensing Letters}, 22:1--5, 2025.
\newblock doi:
  \href{https://doi.org/10.1109/LGRS.2025.3538868}{\nolinkurl{10.1109/LGRS.2025.3538868}}.

\bibitem{zhang2022s2onpde}
Mingjin Zhang, Chengyu He, Jing Zhang, Yuxiang Yang, Xiaoqi Peng, and Jie Guo.
\newblock {SAR}-to-optical image translation via neural partial differential
  equations.
\newblock In {\em Proceedings of the Thirty-First International Joint
  Conference on Artificial Intelligence (IJCAI)}, pages 1644--1650, 2022.
\newblock doi:
  \href{https://doi.org/10.24963/ijcai.2022/229}{\nolinkurl{10.24963/ijcai.2022/229}}.

\bibitem{zhang2024s2otdn}
Mingjin Zhang, Peng Zhang, Yuhan Zhang, Minghai Yang, Xiaofeng Li, Xiaogang
  Dong, and Luchang Yang.
\newblock {SAR}-to-optical image translation via an interpretable network.
\newblock {\em Remote Sensing}, 16(2):242, 2024.
\newblock doi:
  \href{https://doi.org/10.3390/rs16020242}{\nolinkurl{10.3390/rs16020242}}.

\bibitem{bai2024conddiff}
Xinyu Bai, Xinyang Pu, and Feng Xu.
\newblock Conditional diffusion for {SAR} to optical image translation.
\newblock {\em IEEE Geoscience and Remote Sensing Letters}, 21:1--5, 2024.
\newblock doi:
  \href{https://doi.org/10.1109/LGRS.2023.3337143}{\nolinkurl{10.1109/LGRS.2023.3337143}}.

\bibitem{rombach2022high}
Robin Rombach, Andreas Blattmann, Dominik Lorenz, Patrick Esser, and Bj{\"o}rn
  Ommer.
\newblock High-resolution image synthesis with latent diffusion models.
\newblock In {\em Proceedings of the IEEE/CVF Conference on Computer Vision and
  Pattern Recognition (CVPR)}, pages 10684--10695, 2022.
\newblock doi:
  \href{https://doi.org/10.1109/CVPR52688.2022.01042}{\nolinkurl{10.1109/CVPR52688.2022.01042}}.

\bibitem{qin2025e3diff}
Jiang Qin, Bin Zou, Haolin Li, and Lamei Zhang.
\newblock Efficient end-to-end diffusion model for one-step {SAR}-to-optical
  translation.
\newblock {\em IEEE Geoscience and Remote Sensing Letters}, 22:4000405, 2025.
\newblock doi:
  \href{https://doi.org/10.1109/LGRS.2024.3506566}{\nolinkurl{10.1109/LGRS.2024.3506566}}.

\bibitem{espinosa2026copgen}
Miguel Espinosa, Eva Gmelich~Meijling, Valerio Marsocci, Elliot~J. Crowley, and
  Mikolaj Czerkawski.
\newblock {COP-GEN}: Latent diffusion transformer for copernicus earth
  observation data.
\newblock arXiv preprint arXiv:2603.03239, 2026.
\newblock doi:
  \href{https://doi.org/10.48550/arXiv.2603.03239}{\nolinkurl{10.48550/arXiv.2603.03239}}.

\bibitem{espinosa2025copgenbeta}
Miguel Espinosa, Valerio Marsocci, Yuru Jia, Elliot~J. Crowley, and Mikolaj
  Czerkawski.
\newblock {COP-GEN-Beta}: Unified generative modelling of {COPernicus} imagery
  thumbnails.
\newblock In {\em Proceedings of the IEEE/CVF Conference on Computer Vision and
  Pattern Recognition Workshops (CVPRW)}, pages 3110--3120, 2025.
\newblock doi:
  \href{https://doi.org/10.48550/arXiv.2504.08548}{\nolinkurl{10.48550/arXiv.2504.08548}}.

\bibitem{schmitt2019sen12ms}
Michael Schmitt, Lloyd~Haydn Hughes, Chunping Qiu, and Xiao~Xiang Zhu.
\newblock {SEN12MS} -- a curated dataset of georeferenced multi-spectral
  {Sentinel-1/2} imagery for deep learning and data fusion.
\newblock {\em ISPRS Annals of the Photogrammetry, Remote Sensing and Spatial
  Information Sciences}, IV-2/W7:153--160, 2019.
\newblock doi:
  \href{https://doi.org/10.5194/isprs-annals-IV-2-W7-153-2019}{\nolinkurl{10.5194/isprs-annals-IV-2-W7-153-2019}}.

\bibitem{canny1986computational}
John Canny.
\newblock A computational approach to edge detection.
\newblock {\em IEEE Transactions on Pattern Analysis and Machine Intelligence},
  8(6):679--698, 1986.
\newblock doi:
  \href{https://doi.org/10.1109/TPAMI.1986.4767851}{\nolinkurl{10.1109/TPAMI.1986.4767851}}.

\bibitem{oquab2024dinov2}
Maxime Oquab, Timoth{\'e}e Darcet, Th{\'e}o Moutakanni, Huy~V. Vo, Marc
  Szafraniec, Vasil Khalidov, Pierre Fernandez, Daniel Haziza, Francisco Massa,
  Alaaeldin El-Nouby, et~al.
\newblock {DINOv2}: Learning robust visual features without supervision.
\newblock {\em Transactions on Machine Learning Research}, 2024.
\newblock doi:
  \href{https://doi.org/10.48550/arXiv.2304.07193}{\nolinkurl{10.48550/arXiv.2304.07193}}.

\bibitem{lowe2004distinctive}
David~G. Lowe.
\newblock Distinctive image features from scale-invariant keypoints.
\newblock {\em International Journal of Computer Vision}, 60(2):91--110, 2004.
\newblock doi:
  \href{https://doi.org/10.1023/B:VISI.0000029664.99615.94}{\nolinkurl{10.1023/B:VISI.0000029664.99615.94}}.

\end{thebibliography}

\end{document}